\documentclass[10pt,conference,a4paper]{IEEEtran}
\IEEEoverridecommandlockouts
\usepackage{epsfig}
\usepackage{graphicx}
\usepackage{amsmath}
\usepackage{amssymb}
\usepackage{multirow}
\usepackage{multicol}
\graphicspath{./Images}
\usepackage{cite}
\usepackage{amsmath,amssymb,amsfonts}
\usepackage{algorithmic}
\usepackage{graphicx}
\usepackage{textcomp}
\usepackage{xcolor}
\usepackage{comment}
\definecolor{RawSienna}{cmyk}{0,0.72,1,0.45}

\DeclareMathOperator{\E}{\mathbb{E}}
\def\BibTeX{{\rm B\kern-.05em{\sc i\kern-.025em b}\kern-.08em
    T\kern-.1667em\lower.7ex\hbox{E}\kern-.125emX}}
\begin{document}

\title{Attack Agnostic Adversarial Defense via Visual Imperceptible Bound}

\author{
\IEEEauthorblockN{Saheb Chhabra$^{1}$, Akshay Agarwal$^{1,2}$, Richa Singh$^{3}$, and Mayank Vatsa$^{3}$}
\IEEEauthorblockA{$^{1}$IIIT-Delhi, India; $^{2}$Texas A\&M University, Kingsville, USA; $^{3}$IIT Jodhpur, India\\ $^{1}$sahebc@iiitd.ac.in; $^{2}$Akshay.Agarwal@tamuk.edu; $^{3}$\{richa, mvatsa\}@iitj.ac.in
}
}

\maketitle

\begin{abstract}

The high susceptibility of deep learning algorithms against structured and unstructured perturbations has motivated the development of efficient adversarial defense algorithms. However, the lack of generalizability of existing defense algorithms and the high variability in the performance of the attack algorithms for different databases raises several questions on the effectiveness of the defense algorithms. In this research, we aim to design a defense model that is robust within a certain bound against both seen and unseen adversarial attacks. This bound is related to the visual appearance of an image, and we termed it as \textit{Visual Imperceptible Bound (VIB)}. To compute this bound, we propose a novel method that uses the database characteristics. The VIB is further used to measure the effectiveness of attack algorithms. The performance of the proposed defense model is evaluated on the MNIST, CIFAR-10, and Tiny ImageNet databases on multiple attacks that include C\&W ($l_2$) and DeepFool. The proposed defense model is not only able to increase the robustness against several attacks but also retain or improve the classification accuracy on an original clean test set. The proposed algorithm is attack agnostic, i.e. it does not require any knowledge of the attack algorithm.

\end{abstract}

\section{Introduction}

Despite the success of artificial intelligence (AI) systems, the robustness and generalizability against adversarial attacks is still an open question.  Recently, several defense techniques have been proposed by researchers to improve the robustness of AI systems against the adversarial attacks \cite{goel2018smartbox,agarwal2020noise}. 
One of the most effective defense algorithms in the literature is based on the retraining of the target Deep Neural Network (DNN) models using the adversarial examples. However, retraining with specific perturbations may not provide robustness to unseen attacks. 

\begin{figure}[t]
\centering
\includegraphics[scale = 0.325]{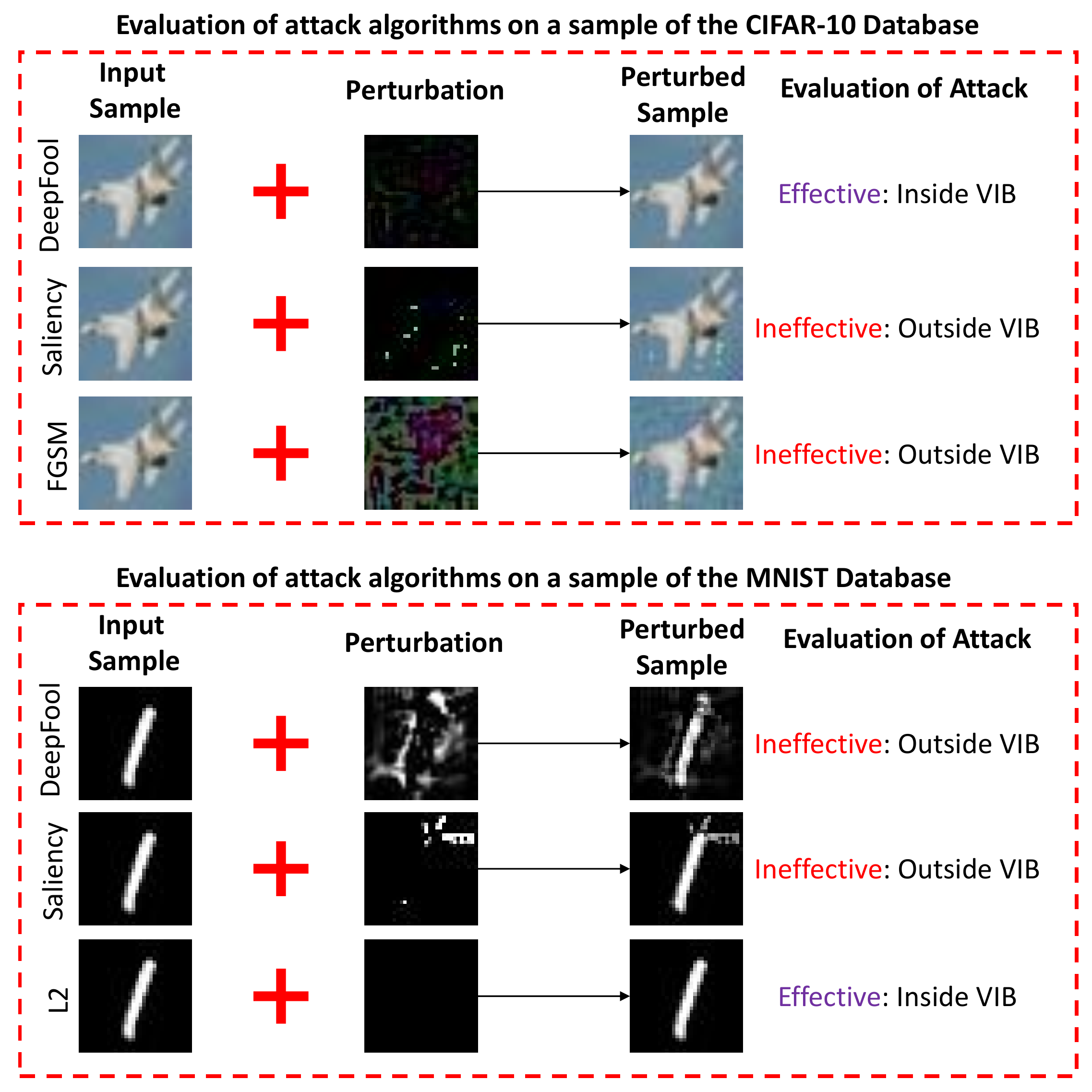}

\caption{Illustrating the performance of the attack on a sample of the CIFAR-10 and MNIST database. In case of the CIFAR-10 database, only DeepFool attack is effective while in case  of the MNIST database, only L2 attack is effective. The effectiveness of an attack algorithm is computed based on Visual Imperceptible bound (VIB).}
\label{fig:Visual_Abstract}
\end{figure}

In order to design an efficient defense model, it is important to first understand the behavior of attack algorithms for different databases and define a certain bound within which the defense model should be robust against both seen and unseen attacks. This bound is related to the bound within which the visual appearance of the image is preserved while performing adversarial manipulation and we termed this bound as \textbf{Visual Imperceptible Bound (VIB)}. It is interesting to observe that the attack algorithms are not equally effective towards all the databases. As shown in Figure \ref{fig:Visual_Abstract}, Deepfool attack is considered to be effective on the CIFAR10 database but ineffective on the MNIST database. Here, the effectiveness of an attack algorithm is defined in terms of the visual imperceptibility of the perturbation noise added in the image. The attack algorithm for a given image is considered to be effective if it can misclassify the given input image by adding a perturbation noise (which results in a perturbed image or adversarial image generated near an original sample) such that the perturbation noise is imperceptible. In other words, the attack algorithm for a given input image is effective if the class of an input image is not equal to the class of an adversarial image and the magnitude of the perturbation is less than $\sigma$. $\sigma$ represents the \textbf{Visual Imperceptible Bound (VIB)} within which the perturbation noise is imperceptible. To measure the visual imperceptible bound of each image for a given database, we propose a novel method that takes into account the given database characteristics and computes a visual imperceptible bound of each image. Using this bound, we also provide a measure to evaluate the effectiveness of an attack algorithm which outputs the score between 0 to 100 where score 0 indicates ``ineffective adversarial attack'' and score 100 indicates ``highly effective adversarial attack''. 

As mentioned above, the aim is to design an efficient defense model such that it is robust within the visual imperceptible bound against both seen and unseen adversarial attacks. For this purpose, we propose a defense algorithm that maps the input samples present within a visual imperceptible bound to the same output. This results in the defense model robust within the VIB against both seen and unseen attacks. The proposed algorithm is attack agnostic, i.e. it does not require any knowledge of the attack algorithm.

\section{Related Work}

With the introduction of the sensitivity of deep learning algorithms towards adversarial perturbation, several defense algorithms have been proposed in the literature. The adversarial defense algorithms are broadly categorized into external classification based defenses, input and model transformations, and certified defenses, and retraining of the target model. 

External classifier based defenses primarily learn a binary classifier along with a hand-crafted or learned feature, or a combination of both  \cite{agarwal2018robustness,goswami2019unravelling,liu2019detection,lu2017safetynet}. Input and model transformation based algorithms transform the images into certain domains and change the network architecture to mitigate the effect of adversarial perturbations \cite{goswami2018unravelling,xie2017mitigating,liu2018towards,xu2017feature}. Agarwal et al. \cite{9207872} proposed a generalized detection algorithm utilizing the fusion of two image transformations. Goel et al. \cite{goel2019robustness,goel2019robustness2,goel2020robustness} proposed the adversarial robustness using cryptography measures to convert the internal layers of CNN into blocks of the blockchain. Researchers are also working towards providing certified defense against adversarial perturbations. However, most of them are evaluated on the first-order adversary or gray-scale images \cite{wong2018scaling,raghunathan2018semidefinite}. Cohen et al. \cite{cohen2019certified} have shown the certified robustness of large scale ImageNet images and proved slight robustness of $12$\% for $l_2$ (bound of radius $3$) attacks.

Among these existing defenses algorithms, retraining based security algorithms have been considered the most robust defense. The successful retraining based security has achieved by augmenting the adversarial images in the training set itself. Goodfellow et al. \cite{goodfellow6572explaining} demonstrated the adversarial robustness of a CNN model by training it on the clean and fast gradient sign (FGSM) generated adversarial images. Later, Madry et al. \cite{madry2017towards} proposed the first-order adversarial attack termed as PGD and claimed that if the model is robust against such attacks, then it can be robust against first-order attacks. However, both these methods are still vulnerable to optimization-based attacks and the attacks developed in black-box settings. To tackle these limitations, Tram{\`e}r et al. \cite{tramer2017ensemble} generated adversarial examples from multiple pre-trained models and used them for retraining the target CNN. Recently, Shafahi et al. \cite{shafahi2019adversarial} proposed the method to improve the adversarial training method by utilizing the network gradient simultaneously for parameter update and adversarial example generation.

Yuan et al. \cite{8611298}, Ren et al. \cite{ren2020adversarial}, and Singh et al. \cite{singh2020robustness} summarized existing adversarial defense algorithms in their recent survey papers. With the increasing usage of AI algorithms, the importance of adversarial learning is also increasing. However, there is a significant scope of improving the understanding and effectiveness of the attacks and the defense algorithms. Except adversarial training, most of the existing defense algorithms are ineffective against optimization-based attacks \cite{athalye2018obfuscated, ghiasi2020breaking}. However, research suggests that adversarial training is vulnerable to black-box attacks with several privacy issues and creates blind-spots for further attacks \cite{zhang2019limitations,mejia2019adversarial}. Recently, Agarwal et al. \cite{agarwal2020sign} have performed a study regarding the essential components in network training and adversarial examples generation, which can further improve adversarial robustness.

\section{Proposed Approach}
Existing techniques such as adversarial training increases the robustness of the model by retraining it with the adversarial examples generated from a particular attack. These techniques result in the model to be robust against the seen attack only. In a real-world scenario, the defense model should be robust to seen as well as unseen attacks. To address this problem, we aim to design a defense model that is robust within a certain bound against both seen and unseen attacks. In this research, this bound is referred to as the bound within which the visual appearance of an image is preserved and we termed this bound as \textbf{Visual Imperceptible Bound (VIB)}. To compute VIB, we propose a novel method that takes into account the database characteristics and computes the bound. Since each image has a different visual imperceptible bound and any adversarial attack is considered successful if perturbed images are generated within the VIB. To design a defense model, we use the concept of visual imperceptible bound and propose a defense algorithm that maps the input samples within the visual imperceptible bound to the same output. This results in the model robust to both seen and unseen attacks within VIB. The details of the method to compute the VIB and the proposed defense algorithm are discussed below. 


Consider a database $\mathbf{D}$ with $m$ number of observations sampled from true data distribution $P(x, y)$, where $x$ represents the clean images and $y$ correspond to the labels encoded in one hot form. Mathematically, the database $D$ is written as:
\begin{equation}
   \mathbf{D} = \{(x_1, y_1), (x_2, y_2),...,(x_m, y_m)\}
\end{equation}

As mentioned above, the aim is to train the defense model that is robust within the visual imperceptible bound. Here, we will first discuss the method to compute the visual imperceptible bound followed by the measure to evaluate the attack algorithm and further, the proposed defense algorithm. 

\subsection{Visual Imperceptible Bound (VIB)} \label{sec:vib}
\begin{figure}[t]
\centering
\includegraphics[scale = 0.4]{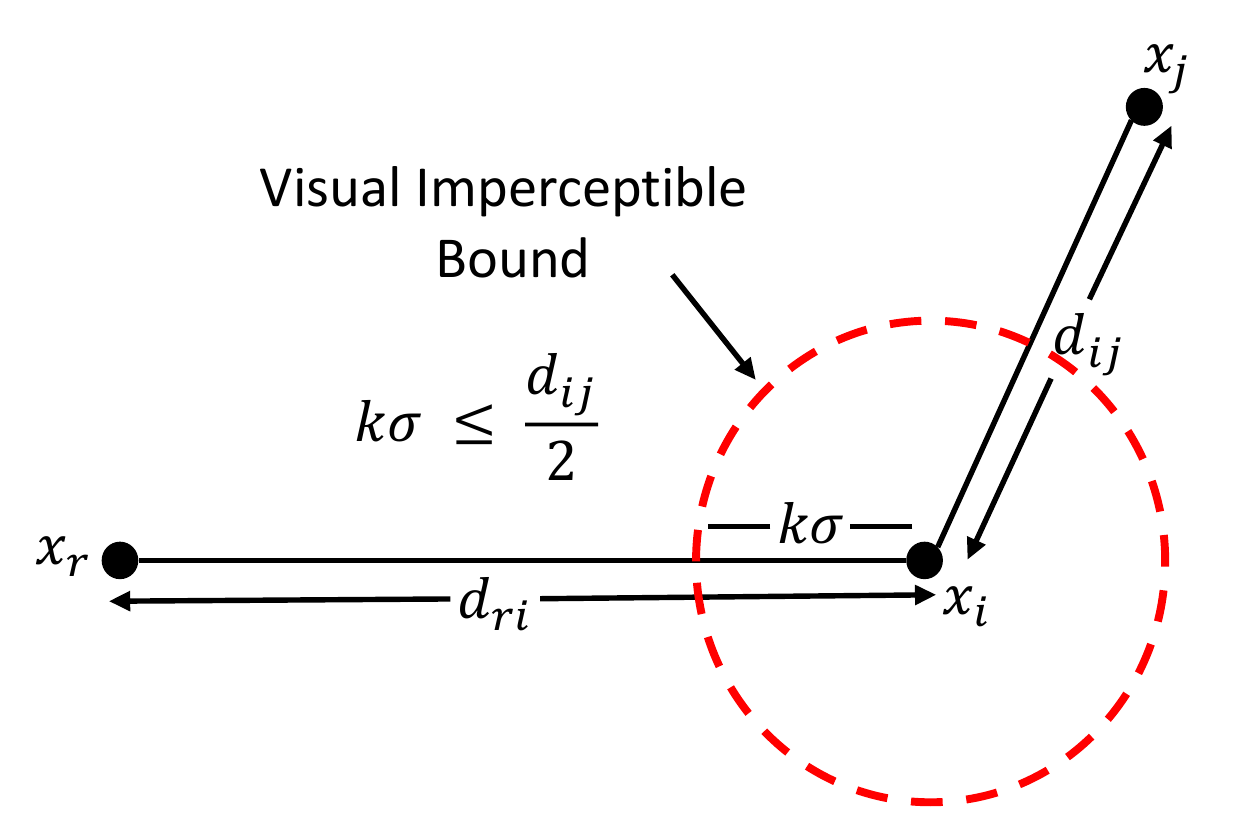}

\caption{Illustrates the method to compute VIB. $x_i$, $x_j$, and $x_p$ are the samples in the input image space. The distance between samples $x_i$ and $x_j$ i.e. $d_{ij}$ is less than the distance between samples $x_p$ and $x_i$ i.e. $d_{pi}$. Therefore, the maximum imperceptible bound of $x_i$ would be $\frac{d_{ij}}{2}$}.
\label{fig:VIB}
\end{figure}

The visual imperceptible bound of a clean image represents the range in which the visual appearance of an image is preserved while performing adversarial manipulation. Let $\sigma_i$ represent the visual imperceptible bound of an image $x_i$. In order to compute the visual imperceptible bound of an image $x_i$, the first step is to compute the nearest neighbor image of the different class for an image $\mathbf{x_i}$. Let $x_j$ represents the nearest neighbor image. In the next step, $l_1$ distance is computed between $x_i$ and $x_j$. Mathematically, it is written as:
\begin{equation}
    d_{ij} = |x_i - x_j|
\end{equation}
where, $d_{ij}$ represents the distance or separation between the images $x_i$ and $x_j$. Here, image $x_i$ and $x_j$ are of different class. Consider the image $x_i$ starts moving towards image $x_j$. After distance $\frac{d_{ij}}{2}$, the visual appearance of the image $x_i$ is dominated by the appearance of the image $x_j$. This means that the upper bound of the visual imperceptible bound (VIB) i.e. $\sigma_i$ should be less than $\frac{d_{ij}}{2}$ as shown in Figure \ref{fig:VIB}. It is well known that for Gaussian distribution 1$\sigma$ covers approximately 68\% region of distribution. Similarly, 2$\sigma$ covers 95\% and 3$\sigma$ covers 99\% region. The relation between visual imperceptible bound $\sigma_i$ and the upper bound of VIB is written as:


\begin{equation}
\begin{split}
    \sigma_i \leq \frac{d_{ij}}{2k}
\end{split}
\end{equation}
where, $k \in \mathbb{N}$.


\subsection{Attack Evaluation}
Once the VIB is computed, the next aim is to measure the effectiveness of the attack algorithm. For this purpose, mean visual imperceptible bound $\sigma_{mean} = \sum_{i=1}^m \sigma_i$ is computed and further, the attack is performed. The percentage of adversarial samples that are present within the mean visual imperceptible bound $\sigma_{mean}$ represents an approximate effectiveness score corresponding to the attack algorithm. The range of effectiveness score is between 0 to 100, where, 0 indicates the ineffective and 100 indicates the highly effective attack. 

\subsection{Defense Algorithm}
Let $f_{\theta_1}$ be the model with parameter $\theta_1$ pre-trained on database $D$ with clean images such that it maps the input $x$ from an input space $\mathbf{\chi}$ to output $y$ which belongs to an output space $\mathbf{\gamma}$. Mathematically, it is written as $f_{\theta_1}: \mathbf{\chi} \rightarrow \mathbf{\gamma}$. For an input image $x_i$, the model $f_{\theta_1}$ outputs the probability vector $p(y_i|x_i)$. The aim is to train the defense model such that it outputs the same prediction for the input samples present within the visual imperceptible bound. In order to ensure that the defense model outputs the same prediction for the samples present within the visual imperceptible bound, we generated the random perturbations from the Gaussian distribution $\mathcal{N}(0, \sigma_i)$. Thus, the objective function for an input image $x_i$ is written as: 
\begin{equation}
    \min_{\theta_2} \hspace{3pt} \E_{(x_i, y_i) \sim \mathbf{D}}  \E_{\triangle p_i \sim \mathcal{N}(0, \sigma_i)} L(f_{\theta_1}(x_i), f_{\theta_2}(x_i + \triangle p_i))
    \label{eq:obj}
\end{equation}


\begin{figure}[t]
\centering
\includegraphics[scale = 0.45]{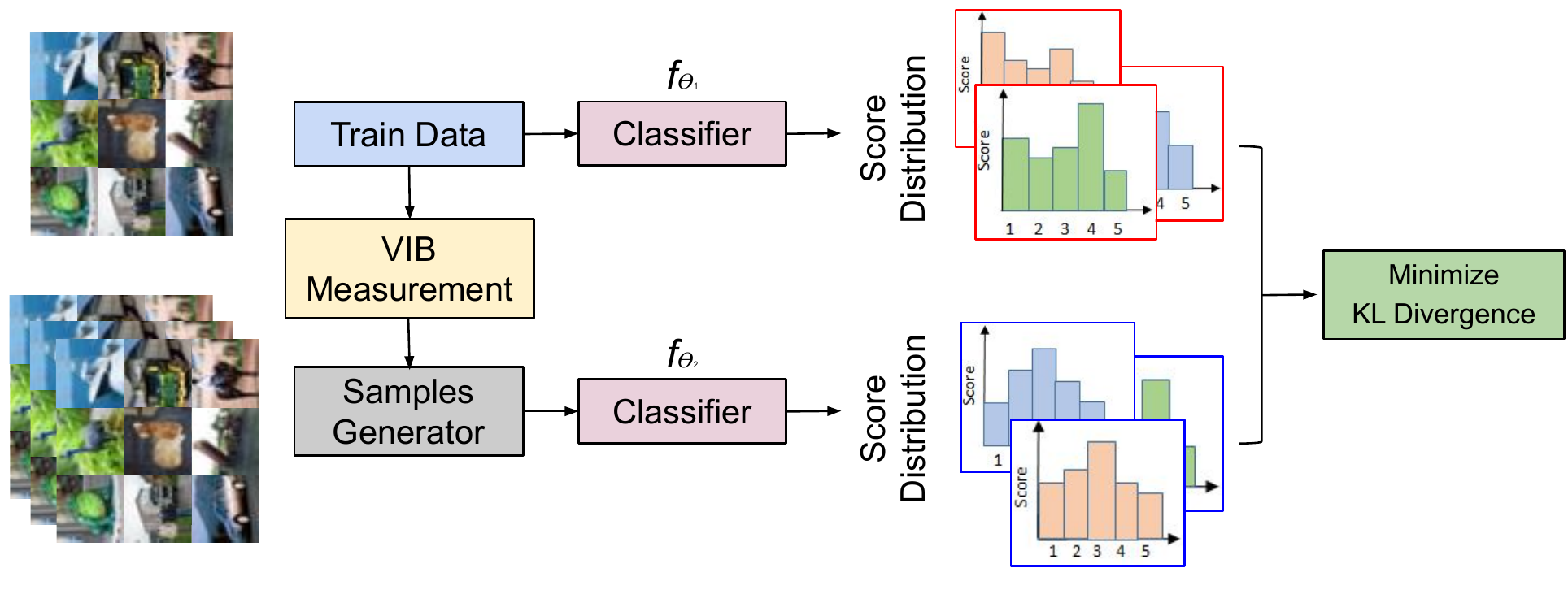}

\caption{Block diagram showing the steps involved in the proposed defense algorithm. In the first step, the output score distribution from the model $f_{\theta_1}$ is computed. In the next step, the model $f_{\theta_2}$ initialized with parameter $\theta_1$ is trained with the data generated from Gaussian distribution. While training, the output score distribution of the samples generated from Gaussian distribution is mapped to the output of model $f_{\theta_1}$.}
\label{fig:Algo}
\end{figure}

where, $\triangle_{p_i}$ represents a random perturbation generated from $\mathcal{N}(0, \sigma_i)$ and $f_{\theta_2}(x_i + \triangle p_i)$ represents the defense model with perturbed sample as an input and model parameters $\theta_2$. $L(.)$ is a function which computes the distance among the output score vectors. The above equation is optimized over model parameter $\theta_2$. It is important to note that while training the defense model $f_{\theta_2}$, the model parameters are initialized with $\theta_1$. Let $n$ be the number of perturbations generated for an input image $x_i$. Thus, Equation \ref{eq:obj} is updated as:
\begin{equation}
    \min_{\theta_2} \hspace{3pt} \E_{(x_i, y_i) \sim \mathbf{D}} \hspace{3pt} \frac{1}{n} \sum_{j = 1}^{n}  L(f_{\theta_1}(x_i), f_{\theta_2}(x_i + \triangle p_{i,j})) 
\end{equation}

In this research, KL divergence loss is minimized over $f_{\theta_1}(x_i)$ and  $f_{\theta_2}(x_i + \triangle p_{i, j})$, and the above equation is updated as:
\begin{equation}
    \min_{\theta_2} \hspace{3pt} \E_{(x_i, y_i) \sim \mathbf{D}} \hspace{3pt} \frac{1}{n} \sum_{j = 1}^{n}  L_{KL}(f_{\theta_1}(x_i) || f_{\theta_2}(x_i + \triangle p_{i, j})) 
\end{equation}
where, $L_{KL}(.)$ represents the KL divergence loss. For a complete database $D$, the final equation is written as: 
\begin{equation}
    \min_{\theta_2} \hspace{3pt} \sum_{i = 1}^{m} \hspace{3pt} \frac{1}{n} \sum_{j = 1}^{n}  L_{KL}(f_{\theta_1}(x_i) || f_{\theta_2}(x_i + \triangle p_{i, j})) 
\end{equation}

\section{Experimental Setup} \label{sec:exp-setup}

The performance of the proposed defense model and the method to compute VIB is evaluated on three publicly available databases: MNIST \cite{krizhevsky2009learning}, CIFAR-10 \cite{lecun2010mnist}, and Tiny ImageNet \cite{yao2015tiny}. The adversarial examples are generated using four different attack algorithms i.e., DeepFool \cite{moosavi2016deepfool}, Fast Gradient Sign Method (FGSM) \cite{goodfellow6572explaining}, Jacobian Saliency Map (JSMA) \cite{papernot2016limitations}, and C\&W (l$_2$) \cite{carlini2017towards}. The results of the proposed defense model are compared with adversarial training \cite{tramer2017ensemble,kurakin2016adversarial}. Additionally, we have evaluated the proposed defense model against the PGD attack in the white box setting and compared the results with existing algorithms. The details of the databases along with their protocol, attack algorithms, and algorithms used for comparison are discussed below.

\noindent \textbf{Databases:} MNIST \cite{lecun2010mnist} database consist of of $60,000$ training images and $10,000$ testing images of $10$ classes with digits $0$ to $9$. CIFAR-10 \cite{krizhevsky2009learning} and Tiny ImageNet \cite{yao2015tiny} databases are primarily used for object recognition. CIFAR-10 consist of $50,000$ training images and $10,000$ testing images of $32\times{32}$ resolution with $10$ different classes. Tiny ImageNet consists of images of $200$ object classes of resolution $64\times64$. Each class consists of $500$ images in training and $50$ images in testing set. For MNIST and CIFAR-10 databases, pre-defined standard protocol is used to train the $f_{\theta_1}$ model. For Tiny ImageNet database, images of randomly selected 40 classes are used for evaluation. To train model $f_{\theta_2}$ using the proposed defense algorithm, $10$, $15$, and $20$ perturbations are generated within the VIB corresponding to each image in the database.


\noindent \textbf{Attack Algorithms:} In this research, we have considered multiple attacks to validate the strength of the proposed adversarial defense model. The Fast Gradient Sign Method (FGSM) attack proposed by Goodfellow et al. \cite{goodfellow6572explaining} is based on linearizing the loss function corresponding to each image and add the noise found on maximizing the loss constraint to l$_\infty$ norm. The FGSM attack uses the gradient output to perform adversarial manipulations. Moosavi-Dezfooli et al. \cite{moosavi2016deepfool} proposed a greedy algorithm termed as DeepFool, which iteratively computes the small magnitude perturbation imperceptible to the human eye. Papernot et al. \cite{papernot2016limitations} introduced a new class of algorithm to fool deep neural networks. They computed feed-forward derivatives and constructed a Jacobian matrix representing each input pixel's effect on the output vector. The saliency map has been computed to obtain the most effective pixels which require minimal change to fool the model. C\&W ($l_2$) is one of the most vigorous optimization-based attacks which can reduce the classification performance to $0$\% on most of the databases. The adversarial images of each type are generated using the toolbox developed by Nicolae et al. \cite{art2018}.



\begin{table}[]
\small
\centering
\caption{Details of experiments on MNIST, CIFAR10, and Tiny ImageNet to showcase the efficacy of the proposed defense model.}

\label{exp-list}
\begin{tabular}{|l|c|c|}
\hline
\textbf{Experiment}                                                      & \textbf{Database}                                         & \textbf{Attack}                                                     \\ \hline
\begin{tabular}[c]{@{}c@{}}Evaluation of \\ Database\end{tabular} & \begin{tabular}[c]{@{}c@{}}MNIST, CIFAR-10\\ Tiny ImageNet\end{tabular}  & -                                                                   \\ \hline
\begin{tabular}[c]{@{}c@{}}Evaluation of \\ Attacks\end{tabular}                                                        & \begin{tabular}[c]{@{}c@{}}MNIST, CIFAR-10\\ Tiny ImageNet\end{tabular} & \begin{tabular}[c]{@{}c@{}}DeepFool, FGSM, \\ JSMA, $l_2$\end{tabular} \\ \hline
\begin{tabular}[c]{@{}c@{}}Proposed Defense\\  Model\end{tabular}        & \begin{tabular}[c]{@{}c@{}}MNIST, CIFAR-10\\ Tiny imageNet\end{tabular} & \begin{tabular}[c]{@{}c@{}}DeepFool, FGSM, \\ JSMA, $l_2$\end{tabular} \\ \hline
\end{tabular}
\end{table}

\noindent \textbf{Defense Algorithms used for comparison:} The most effective defense algorithms in the literature are based on the adversarial training of the target network. Adversarial training is referred to the retraining of the CNN model where the training data is augmented with the adversarial data. In the adversarial training based defense algorithms, the adversarial examples are computed using the target network, which is not feasible for large-scale databases such as ImageNet. To counter this limitation Kurakin et al. \cite{kurakin2016adversarial} Tram{\'e}r et al. \cite{tramer2017ensemble} and Madry et al. \cite{madry2017towards} have proposed the variant of adversarial training where the target network is retrained with one specific type of adversarial examples computed on another CNN model.

\noindent \textbf{Implementation Details:} To train the model $f_{theta_1}$, a pre-trained VGG16 model is fine-tuned on clean images for 20 epochs with Adam optimizer for MNIST and CIFAR-10 databases. For the Tiny ImageNet database, the model VGG16 is trained from scratch. For training the proposed defense model, the model trained with clean images is further optimized with the new samples obtained using the proposed defense algorithm. The learning rate is set to 0.001 and the new samples are trained for 10 epochs.

\section{Performance Evaluation}
This section summarizes the results and findings of the experiments related to the evaluation of attack algorithms and the proposed defense algorithm. We have performed three kinds of experiments: 1) evaluating the vulnerability of databases, 2) evaluation of attacks and 3) evaluation of the proposed defense model. The first two experiments provide insight into the vulnerability of databases against different attacks within VIB. The third experiment is performed to showcase the performance of the proposed defense model within VIB against attacks. Further, the proposed defense model is also compared with existing algorithms.

\subsection{Evaluation of Vulnerability of Databases}
To evaluate the vulnerability of databases against attacks, the nearest neighbor distance of different class samples is computed. This distance represents the vulnerability of images of a particular class to be misclassified into another class through perturbation. The lesser the distance between the samples of different classes, the more vulnerable the database is to attack. This is one of the important reasons for the sudden change of different CNN model accuracy via changes in image pixel space with respect to different classes. 

\begin{figure}[t]
\centering
\includegraphics[scale = 0.59]{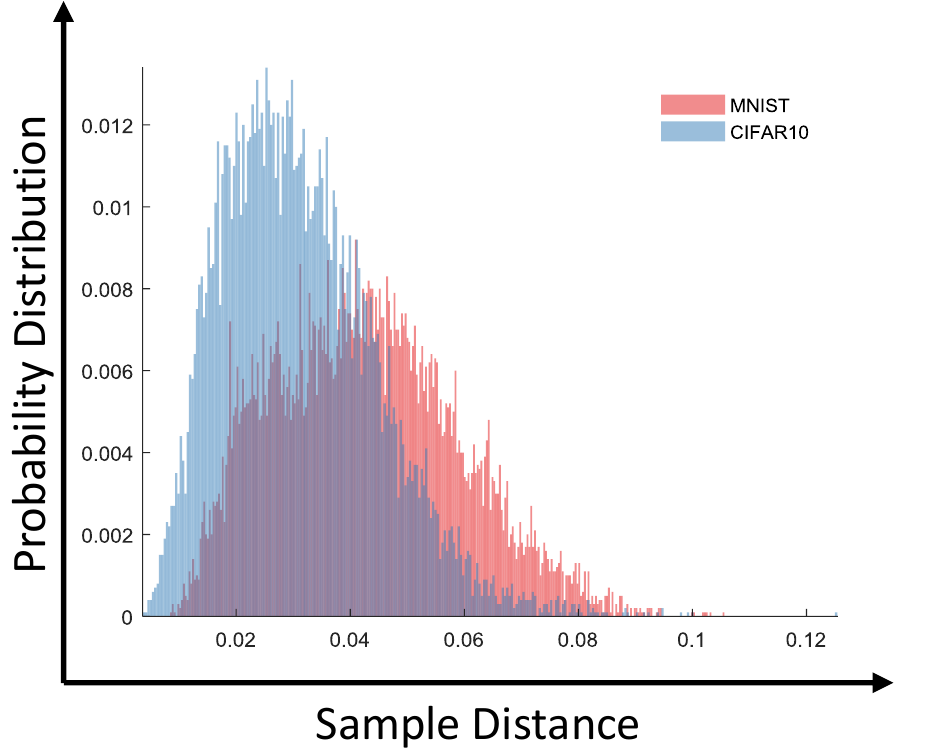}

\caption{Comparison of normalized class distance in image space on the MNIST and CIFAR10 database.}
\label{fig:Comp-img-dist}
\end{figure}

\begin{figure}
\centering
\includegraphics[scale = 0.59]{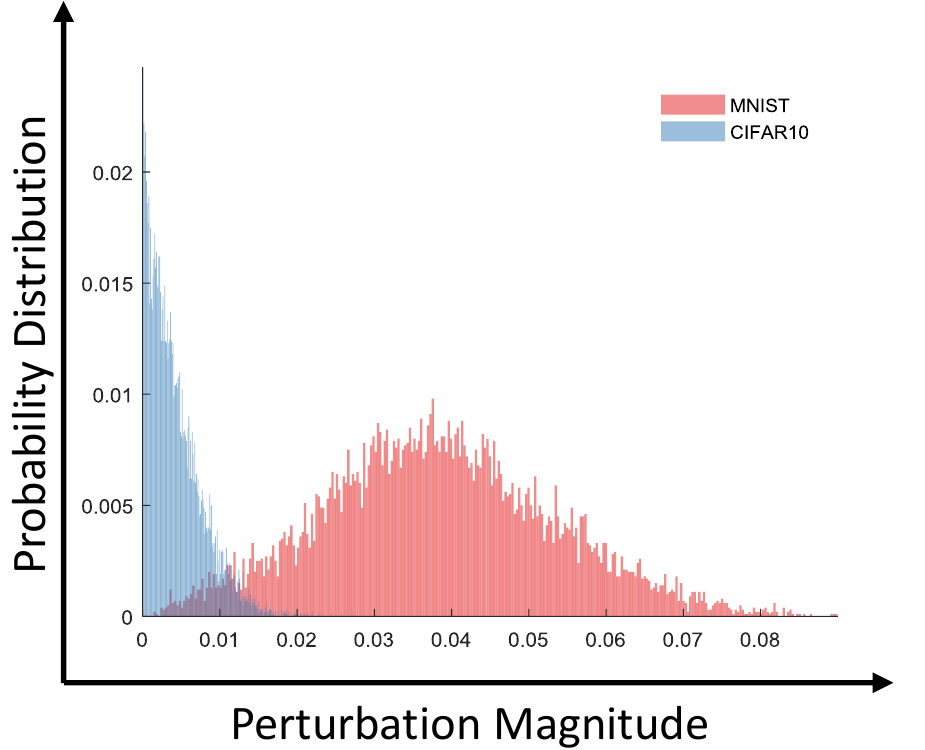}

\caption{Comparison of normalized magnitude of perturbation noise added to perform attack on the MNIST and CIFAR10 database using DeepFool attack algorithm.}
\label{fig:Comp_Pert_Magn}
\end{figure}

The results of the vulnerability of MNIST and CIFAR10 databases are shown in Figure \ref{fig:Comp-img-dist}. It is interesting to note that the distances between the samples of different classes in MNIST are larger in comparison to the CIFAR10 database which makes the perturbation noise perceptible in the image. To evaluate the proposed method, the DeepFool attack is performed on the model $f_{\theta_1}$ to generate adversarial examples which result in a similar drop in accuracy for both the MNIST and CIFAR10 databases, i.e. $34.19\%$ and $31.67\%$. Afterward, the perturbation magnitude for each sample in both the databases is computed. The distribution of the magnitude of perturbation noise for both the databases is shown in Figure \ref{fig:Comp_Pert_Magn}. It is clear from the Figure that the CIFAR10 has less magnitude of perturbation noise as compared to the MNIST to reach a similar drop in accuracy. This shows that the CIFAR10 is more vulnerable to adversarial attacks and hence can be fooled with adversarial noise while maintaining the visual appearance of the images. 


\subsection{Evaluation of Attack Algorithms}
To evaluate the effectiveness of attack algorithms, four different attacks are considered namely 1) DeepFool, 2) FGSM, 3) JSMA, and 4) $L_2$ attack. These attacks are evaluated based on whether the adversarial examples are generated inside the mean VIB or not. For this purpose, adversarial examples are generated for both the databases with the above mentioned attacks. We will first discuss VIB for all the databases followed by the evaluation of attacks.

\noindent \textbf{Visual Imperceptible Bound:} As mentioned in section \ref{sec:vib}, $\sigma$ represents the visual imperceptible bound (VIB) of the image. As each image has a different VIB, therefore, we compute the minimum imperceptible bound of the database, mean imperceptible bound of the database, and maximum imperceptible bound of the database. This represents the best (minimum), average (mean), and worst-case (maximum) for a database to be attacked by the attack algorithms. The results of computation of VIB for $k=1$, $k=2$, $k=2.5$, and $k=3$ for all databases are shown in Table \ref{tab:CIFAR_10_BAW}. On comparing the VIB for both the databases, it is observed that the MNIST database has a lower VIB. This implies that noise would become perceptible even with very less magnitude of the perturbation.

\noindent \textbf{Attack Evaluation:} To evaluate the attack algorithms, we have considered the mean (average case) VIB. As mentioned earlier, the percentage of adversarial examples generated within the mean VIB represents the effectiveness score. The results are shown in Table \ref{tab:Estimated_Accuracy}. It is observed that DeepFool, JSMA, $l_2$, and FGSM with $\epsilon=2$ and $4$ generate most of the samples within the bound on the CIFAR-10 database. This shows that these attacks are effective in terms of preserving the visual appearance of the image. For the MNIST database, the adversarial images generated by all the attacks are outside the bound. It shows that all the attacks are ineffective in terms of preserving the visual appearance of the image on the MNIST database. The percentage of adversarial examples within the VIB is expected to be defended by the proposed defense model. Therefore, the percentage of adversarial samples within the VIB can be considered as the estimated accuracy of the proposed defense model.
\begin{table}[t]
\centering
\caption{Minimum (Best), Mean (Average), and Maximum (Worst) $\sigma$ on CIFAR 10, MNIST, and Tiny ImageNet databases.}

\label{tab:CIFAR_10_BAW}
\begin{tabular}{|l|c|c|c|c|}
\hline
\multirow{2}{*}{\textbf{Databases}} & & \textbf{Minimum}  & \textbf{Mean}     & \textbf{Maximum}  \\ \cline{2-5}
                                  &                & \textbf{$\sigma$} & \textbf{$\sigma$} & \textbf{$\sigma$} \\ \hline
\multirow{4}{*}{\textbf{CIFAR10}} & \textbf{k=1}   & 0.011             & 0.062             & 0.127             \\ \cline{2-5}
                                  & \textbf{k=2}   & 0.005             & 0.031             & 0.063             \\ \cline{2-5}
                                  & \textbf{k=2.5} & 0.004             & 0.025             & 0.051             \\ \cline{2-5}
                                  & \textbf{k = 3} & 0.003             & 0.020             & 0.042             \\ \hline
\multirow{4}{*}{\textbf{MNIST}}   & \textbf{k=1}   & 0.003             & 0.023             & 0.065             \\ \cline{2-5}
                                  & \textbf{k=2}   & 0.002             & 0.012             & 0.033             \\ \cline{2-5}
                                  & \textbf{k=2.5} & 0.002             & 0.009             & 0.026             \\ \cline{2-5}
                                  & \textbf{k=3}   & 0.001             & 0.007             & 0.021             \\ \hline
                            
\multirow{4}{*}{\textbf{TinyImageNet}}   & \textbf{k=1}   & 0.017             & 0.076             & 0.1485             \\ \cline{2-5}
                                  & \textbf{k=2}   & 0.009             & 0.038             & 0.074             \\ \cline{2-5}
                                  & \textbf{k=2.5} & 0.007             & 0.030             & 0.059             \\ \cline{2-5}
                                  & \textbf{k=3}   & 0.006             & 0.025             & 0.049             \\ \hline
\end{tabular}
\end{table}

\begin{table}[t]
\centering
\small
\caption{Estimated classification accuracy on attack algorithms within the imperceptible bound with $k = 2$ measured using nearest neighbors distance.}

\label{tab:Estimated_Accuracy}
\begin{tabular}{|c|c|c|c|c|}
\hline
\multirow{8}{*}{\textbf{CIFAR10}} & \textbf{Attack}                                                        & DeepFool                                                         & JSMA                                                             & \begin{tabular}[c]{@{}c@{}}FGSM \\ $\epsilon = 2$\end{tabular}   \\ \cline{2-5}
                                  & \textbf{\begin{tabular}[c]{@{}c@{}}Estimated\\ Accuracy\end{tabular}}  & 83.9                                                             & 83.91                                                            & 83.91                                                            \\ \cline{2-5}
                                  & \textbf{Attack}                                                        & \begin{tabular}[c]{@{}c@{}}FGSM \\ $\epsilon = 4$\end{tabular}   & \begin{tabular}[c]{@{}c@{}}FGSM \\ $\epsilon = 8$\end{tabular}   & \begin{tabular}[c]{@{}c@{}}$l_2$\end{tabular}  \\ \cline{2-5}
                                  & \textbf{\begin{tabular}[c]{@{}c@{}}Estimated \\ Accuracy\end{tabular}} & 83.91                                                            & 18.15                                                            & 83.91                                                                \\ \hline
\multirow{8}{*}{\textbf{MNIST}}   & \textbf{Attack}                                                                 & DeepFool                                                         & JSMA                                                             & \begin{tabular}[c]{@{}c@{}}FGSM \\ $\epsilon = 0.1$\end{tabular} \\ \cline{2-5}
                                  & \textbf{\begin{tabular}[c]{@{}c@{}}Estimated\\ Accuracy\end{tabular}}                                                              &  3.48                                                                & 32.02                                                                 & 0                                                                  \\ \cline{2-5}
                                  & \textbf{Attack}                                                                 & \begin{tabular}[c]{@{}c@{}}FGSM \\ $\epsilon = 0.2$\end{tabular} & \begin{tabular}[c]{@{}c@{}}FGSM \\ $\epsilon = 0.3$\end{tabular} &  $l_2$                                                                 \\ \cline{2-5}
                                  & \textbf{\begin{tabular}[c]{@{}c@{}}Estimated\\ Accuracy\end{tabular}}                                                               & 0                                                             & 0                                                           & 17.85                                                                 \\ \hline
\end{tabular}
\end{table}

\subsection{Proposed Defense Model}

To evaluate the model trained using the proposed defense algorithm, adversarial examples generated using existing attack algorithms are evaluated on the original model $f_{\theta_1}$ and the proposed defense model $f_{\theta_2}$ in the black-box setting. 
For training the proposed defense model, $10$, $15$, and $20$ perturbations are generated corresponding to each image. The experiments are performed for $k = 1, 2, 2.5$, and $3$ which results in $\sigma_i=\frac{d_{ij}}{2}, \frac{d_{ij}}{4}, \frac{d_{ij}}{5}$, and $\frac{d_{ij}}{6}$ where, $d_{ij}$ represents the distance between two nearest images $i$ and $j$ among different classes. Tables \ref{resuts-mnist-prop}, \ref{resuts-cifar10-prop}, and \ref{resuts-tiny-prop} show the performance of the original and proposed model on the test set of clean and adversarial examples of the MNIST and CIFAR-10 respectively.

\noindent \textbf{Results on MNIST:} The classification accuracies of original $f_{\theta_1}$ and the proposed defense model $f_{\theta_2}$ on clean and adversarial examples generated on test set are shown in Table \ref{resuts-mnist-prop}. It is observed that within VIB, the proposed defense model is robust against the adversarial attacks: DeepFool, JSMA, and $l_2$ attack. For instance, the classification accuracy of the original model on adversarial examples generated using the DeepFool and C\&W ($l_2$) attack is $34.19$\% and $12.89$\%. While the proposed defense model outputs the classification accuracy $96.04$\% and $90.63$\% for $k=1$ and $k=2$. This shows the efficacy of the proposed defense model. On comparing the results of the proposed defense model with the estimated accuracy reported in Table \ref{tab:Estimated_Accuracy}, it is observed that that the proposed defense model is able to perform much better. For instance, the estimated accuracy for C\&W attack which is known as one of the stronger optimization based attacks is $17.85$\%. The results obtained using the proposed defense model is $90.63$\% as reported in Table \ref{resuts-mnist-prop}. It is not only greater than the estimated accuracy but also more than the original model's accuracy. This shows that the proposed defense model is adversarially robust against both stronger and weaker attacks.

\noindent \textbf{Results on CIFAR-10:} The results corresponding to clean and adversarial examples generated on the test set are reported in Table \ref{resuts-cifar10-prop}. It is observed that the proposed model is robust against the attacks within VIB, i.e., DeepFool, FGSM with $\epsilon=2$, and JSMA attacks. For instance, after DeepFool attack and FGSM ($\epsilon=2$), the original model outputs $31.67$\% and $55.91$\% accuracy while the classification accuracy of the proposed model is $85.94$\% and $83.53$\% for $k=1$. Similarly, for one of the strongest optimization-based attacks, i.e., the C\&W attack, the proposed defense model is able to regain the object recognition performance. The recognition performance of the proposed robust model under multiple adversarial attack examples showcases the strength of the defense. It is interesting to observe that in some cases such as for C\&W and DeepFool adversarial examples, the accuracy of the proposed defense model is better than original model tested on a clean set. The proposed method is not only able to resist the adversarial attacks but also able to increase the recognition accuracy on the clean test set from $83.91$\% to $88.82$\%.

\begin{table}[t]
\centering
\caption{Classification accuracies (\%) on MNIST database using original and proposed defense algorithm with different number of samples and different values of $k$ ($\sigma = \frac{d_{ij}}{2k}$).}

\label{resuts-mnist-prop}
\setkeys{Gin}{keepaspectratio}
\resizebox*{0.475\textwidth}{0.475\textheight} {
\begin{tabular}{|l|c|c|c|c|c|c|}\hline
\multirow{2}{*}{\textbf{Data Type}}      & \multicolumn{1}{l|}{\multirow{2}{*}{\textbf{Samples}}} & \multicolumn{1}{l|}{\multirow{2}{*}{\begin{tabular}[c|]{@{}l@{}}\textbf{Original} \\ \textbf{Model}\end{tabular}}} & \multicolumn{4}{c|}{\textbf{Proposed Robust Model with}}                                                                         \\ \cline{4-7}
                                & \multicolumn{1}{l|}{}                         & \multicolumn{1}{l}{}                                                                           & \multicolumn{1}{|l|}{\textbf{k=1.0}} & \multicolumn{1}{l|}{\textbf{k=2.0}} & \multicolumn{1}{l|}{\textbf{k=2.5}} & \multicolumn{1}{l|}{\textbf{k=3.0}} \\ \hline
\multirow{3}{*}{Original}       & 10                                           & \multirow{3}{*}{\textbf{99.55}}                                                                         & 99.40                     & 99.48                     & 99.33                     & 99.50                     \\ \cline{4-7} \cline{2-2}
                                & 15                                           &                                                                                                & 99.33                     & 99.52                     & 99.41                     & 99.37                     \\ \cline{4-7} \cline{2-2}
                                & 20                                           &                                                                                                & 99.32                     & 99.17                     & 99.48                     & \textbf{99.53}                     \\ \hline
\multirow{3}{*}{DeepFool}       & 10                                           & \multirow{3}{*}{\textbf{34.19}}                                                                         & 94.13                     & 93.47                     & 95.67                     & 94.30                     \\ \cline{4-7} \cline{2-2}
                                & 15                                           &                                                                                                & 91.79                     & 92.49                     & 95.95                     & 95.65                     \\ \cline{4-7} \cline{2-2}
                                & 20                                           &                                                                                                & \textbf{96.04}                     & 92.67                     & 96.38                     & 95.12                     \\ \hline
\multirow{3}{*}{\begin{tabular}[c]{@{}l@{}}FGSM\\ ($\epsilon=0.1$)\end{tabular}} & 10                                           & \multirow{3}{*}{\textbf{89.65}}                                                                         & 97.91                     & 97.88                     & 97.88                     & 98.14                     \\ \cline{4-7} \cline{2-2}
                                & 15                                           &                                                                                                & 97.55                     & 97.90                     & 98.27                     & 98.28                     \\ \cline{4-7} \cline{2-2}
                                & 20                                           &                                                                                                & 98.22                     & 97.81                     & \textbf{98.36}                     & 98.35                     \\ \hline
\multirow{3}{*}{\begin{tabular}[c]{@{}l@{}}FGSM\\ ($\epsilon=0.2$)\end{tabular}} & 10                                           & \multirow{3}{*}{\textbf{54.52}}                                                                         & 86.21                     & 87.35                     & 89.93                     & 88.20                     \\ \cline{4-7} \cline{2-2}
                                & 15                                           &                                                                                                & 81.43                     & 80.11                     & 89.29                     & \textbf{90.81}                     \\ \cline{4-7} \cline{2-2}
                                & 20                                           &                                                                                                & 89.34                     & 86.48                     & 88.21                     & 84.70                     \\ \hline
\multirow{3}{*}{\begin{tabular}[c]{@{}l@{}}FGSM\\ ($\epsilon=0.3$)\end{tabular}} & 10                                           & \multirow{3}{*}{\textbf{21.82}}                                                                         & 37.60                     & 49.12                     & 48.77                     & \textbf{54.30}                      \\ \cline{4-7} \cline{2-2}
                                & 15                                           &                                                                                                & 36.74                     & 33.78                     & 52.86                     & 51.11                     \\ \cline{4-7} \cline{2-2}
                                & 20                                           &                                                                                                & 38.32                     & 38.53                     & 47.64                     & 34.28                     \\ \hline
\multirow{3}{*}{JSMA}           & 10                                           & \multirow{3}{*}{\textbf{0.10}}                                                                          & 71.75                     & 64.30                     & 77.86                     & 78.02                     \\ \cline{4-7} \cline{2-2}
                                & 15                                           &                                                                                                & 68.13                     & 58.89                     & \textbf{80.36}                     & 60.94                     \\ \cline{4-7} \cline{2-2}
                                & 20                                           &                                                                                                & 73.92                     & 62.81                     & 71.94                     & 63.13              \\ \hline

\multirow{3}{*}{C\&W ($l_2$)}           & 10                                           & \multirow{3}{*}{\textbf{12.89}}                                                                          & 87.98                     & 87.92                     & 88.47                     & 84.90                     \\ \cline{4-7} \cline{2-2}
                                & 15                                           &                                                                                                & 86.18                     & \textbf{90.63}                     & 88.32                     & 88.97                     \\ \cline{4-7} \cline{2-2}
                                & 20                                           &                                                                                                & 87.97                     & 84.87                     & 91.15                     & 91.43              \\ \hline
\end{tabular}
}
\end{table}

\begin{table}[t]
\centering
\caption{Classification accuracies (\%) on CIFAR-10 database using original and proposed defense algorithm with different number of samples and different values of $k$ ($\sigma = \frac{d_{ij}}{2k}$).}

\label{resuts-cifar10-prop}
\setkeys{Gin}{keepaspectratio}
\resizebox*{0.475\textwidth}{0.475\textheight} {
\begin{tabular}{|l|c|c|c|c|c|c|}\hline
\multirow{2}{*}{\textbf{Data Type}}      & \multicolumn{1}{l|}{\multirow{2}{*}{\textbf{Samples}}} & \multicolumn{1}{l|}{\multirow{2}{*}{\begin{tabular}[c|]{@{}l@{}}\textbf{Original} \\ \textbf{Model}\end{tabular}}} & \multicolumn{4}{c|}{\textbf{Proposed Robust Model with}}                                                                         \\ \cline{4-7}
                                & \multicolumn{1}{l|}{}                         & \multicolumn{1}{l}{}                                                                           & \multicolumn{1}{|l|}{\textbf{k=1.0}} & \multicolumn{1}{l|}{\textbf{k=2.0}} & \multicolumn{1}{l|}{\textbf{k=2.5}} & \multicolumn{1}{l|}{\textbf{k=3.0}} \\ \hline
\multirow{3}{*}{Original}       & 10                                           & \multirow{3}{*}{\textbf{83.91}}                                                                         & 87.52                     & 87.59                     & 87.27                     & 86.58                     \\ \cline{4-7} \cline{2-2}
                                & 15                                           &                                                                                                & 88.05                     & 87.94                     & 88.37                     & 88.45                     \\ \cline{4-7} \cline{2-2}
                                & 20                                           &                                                                                                & 88.41                     & \textbf{88.82}                     & 88.64                     & 88.11                     \\ \hline
\multirow{3}{*}{DeepFool}       & 10                                           & \multirow{3}{*}{\textbf{31.67}}                                                                         & 83.64                     & 83.75                     & 82.83                     & 81.95                     \\ \cline{4-7} \cline{2-2}
                                & 15                                           &                                                                                                & 84.26                     & 84.71                     & 85.78                     & 85.58                     \\ \cline{4-7} \cline{2-2}
                                & 20                                           &                                                                                                & \textbf{85.94}                     & 85.59                     & 85.43                     & 85.21                     \\ \hline
\multirow{3}{*}{\begin{tabular}[c]{@{}l@{}}FGSM\\ ($\epsilon=2.0$)\end{tabular}} & 10                                           & \multirow{3}{*}{\textbf{55.91}}                                                                         & 81.32                     & 81.32                     & 80.35                     & 79.71                     \\ \cline{4-7} \cline{2-2}
                                & 15                                           &                                                                                                & 81.96                     & 82.56                     & 83.06                     & 82.70                     \\ \cline{4-7} \cline{2-2}
                                & 20                                           &                                                                                                & \textbf{83.53}                     & 83.11                     & 82.51                     & 82.91                     \\ \hline
\multirow{3}{*}{\begin{tabular}[c]{@{}l@{}}FGSM\\ ($\epsilon=4.0$)\end{tabular}} & 10                                           & \multirow{3}{*}{\textbf{30.51}}                                                                         & 67.29                     & 67.38                     & 66.77                     & 66.27                     \\ \cline{4-7} \cline{2-2}
                                & 15                                           &                                                                                                & 67.70                     & 69.15                     & 69.63                     & 69.84                     \\ \cline{4-7} \cline{2-2}
                                & 20                                           &                                                                                                & \textbf{70.92}                     & 70.60                     & 69.95                     & 69.73                     \\ \hline
\multirow{3}{*}{\begin{tabular}[c]{@{}l@{}}FGSM\\ ($\epsilon=8.0$)\end{tabular}} & 10                                           & \multirow{3}{*}{\textbf{15.76}}                                                                         & 42.05                     & 41.92                     & 41.30                     & 41.11                     \\ \cline{4-7} \cline{2-2}
                                & 15                                           &                                                                                                & 41.42                     & 43.41                     & 43.51                     & 43.58                     \\ \cline{4-7} \cline{2-2}
                                & 20                                           &                                                                                                & \textbf{44.74}                     & 44.59                     & 44.44                     & 44.21                     \\ \hline
\multirow{3}{*}{\begin{tabular}[c]{@{}l@{}}FGSM\\ ($\epsilon=16.0$)\end{tabular}}  & 10                                           & \multirow{3}{*}{\textbf{11.13}}                                                                         & 21.75                     & 21.38                     & 21.52                     & 21.26                     \\ \cline{4-7} \cline{2-2}
                                & 15                                           &                                                                                                & 20.72                     & 22.32                     & 22.14                     & 22.76                     \\ \cline{4-7} \cline{2-2}
                                & 20                                           &                                                                                                & \textbf{22.80}                     & 22.59                     & 21.98                     & 22.74                     \\ \hline
\multirow{3}{*}{JSMA}           & 10                                           & \multirow{3}{*}{\textbf{1.14}}                                                                          & 70.20                     & 70.48                     & 69.99                     & 68.58                     \\ \cline{4-7} \cline{2-2}
                                & 15                                           &                                                                                                & 70.02                     & 72.05                     & 71.13                     & \textbf{73.88}                     \\ \cline{4-7} \cline{2-2}
                                & 20                                           &                                                                                                & 73.01                     & 72.65                     & 72.31                     & 73.41               \\ \hline

\multirow{3}{*}{C\&W ($l_2$)}           & 10                                           & \multirow{3}{*}{\textbf{12.10}}                                                                          &  83.70 & 83.84 & 82.81 & 82.17  \\ \cline{4-7} \cline{2-2}
                                & 15                                           &                         &
84.32 & 84.85 & 85.90 & 85.46          \\ \cline{4-7} \cline{2-2}
                                & 20                                           &                         &
\textbf{86.03} & 85.72 & 85.56 & 85.32              \\ \hline
\end{tabular}
}
\end{table}


\begin{table}[]
\centering
\caption{Classification accuracies (\%) on the Tiny ImageNet database using original and proposed defense algorithm with different number of samples and different values of $k$ ($\sigma = \frac{d_{ij}}{2k}$).}

\label{resuts-tiny-prop}
\setkeys{Gin}{keepaspectratio}
\resizebox*{0.475\textwidth}{0.475\textheight} {
\begin{tabular}{|l|c|c|c|c|c|c|}
\hline
\multirow{2}{*}{\textbf{Data Type}} & \multirow{2}{*}{\textbf{Samples}} & \multirow{2}{*}{\begin{tabular}[c]{@{}c@{}}Original\\ Model\end{tabular}} & \multicolumn{4}{c|}{\textbf{Proposed Robust Model With}}                   \\ \cline{4-7} 
                                    &                                   &                                                                           & \textbf{k=1.0} & \textbf{k=2.0} & \textbf{k=2.5} & \textbf{k=3.0} \\ \hline
\multirow{3}{*}{Original}           & 10                                & \multirow{3}{*}{\textbf{66.60}}                                           & 63.30          & 62.75          & 61.40          & 61.60          \\ \cline{2-2} \cline{4-7} 
                                    & 15                                &                                                                           & 62.65          & 62.40          & 60.50          & 59.55          \\ \cline{2-2} \cline{4-7} 
                                    & 20                                &                                                                           & 61.75          & \textbf{63.40} & 59.15          & 60.95          \\ \hline
\multirow{3}{*}{Deepfool}           & 10                                & \multirow{3}{*}{\textbf{23.60}}                                           & 51.85          & 50.80          & 50.10          & 49.30          \\ \cline{2-2} \cline{4-7} 
                                    & 15                                &                                                                           & 51.90          & 51.65          & 47.70          & 48.95          \\ \cline{2-2} \cline{4-7} 
                                    & 20                                &                                                                           & 50.20          & \textbf{52.00} & 48.35          & 51.40          \\ \hline
\multirow{3}{*}{\begin{tabular}[c]{@{}l@{}}FGSM\\ ($\epsilon=16.0$)\end{tabular}} & 10                                & \multirow{3}{*}{\textbf{38.50}}                                           & 48.50          & 49.25          & 47.45          & 48.85          \\ \cline{2-2} \cline{4-7} 
                                    & 15                                &                                                                           & 50.45          & 48.35          & 46.05          & 47.75          \\ \cline{2-2} \cline{4-7} 
                                    & 20                                &                                                                           & 49.15          & 48.65          & 47.70          & \textbf{49.50} \\ \hline
\multirow{3}{*}{\begin{tabular}[c]{@{}l@{}}FGSM\\ ($\epsilon=4.0$)\end{tabular}}  & 10                                & \multirow{3}{*}{\textbf{32.55}}                                           & 37.35          & 37.50          & 38.10          & 38.10          \\ \cline{2-2} \cline{4-7} 
                                    & 15                                &                                                                           & 38.50          & 38.00          & 36.95          & 38.20          \\ \cline{2-2} \cline{4-7} 
                                    & 20                                &                                                                           & 39.40          & 38.95          & 37.95          & \textbf{39.75} \\ \hline
\multirow{3}{*}{\begin{tabular}[c]{@{}l@{}}FGSM\\ ($\epsilon=8.0$)\end{tabular}}  & 10                                & \multirow{3}{*}{\textbf{30.15}}                                           & 31.35          & 31.75          & 31.70          & 31.75          \\ \cline{2-2} \cline{4-7} 
                                    & 15                                &                                                                           & 32.10          & 31.65          & 31.65          & 31.95          \\ \cline{2-2} \cline{4-7} 
                                    & 20                                &                                                                           & \textbf{32.85} & 32.40          & 31.50          & 32.60          \\ \hline
\multirow{3}{*}{\begin{tabular}[c]{@{}l@{}}FGSM\\ ($\epsilon=16.0$)\end{tabular}} & 10                                & \multirow{3}{*}{\textbf{25.45}}                                           & 24.85          & 25.95          & 26.50          & \textbf{26.60} \\ \cline{2-2} \cline{4-7} 
                                    & 15                                &                                                                           & 26.45          & 25.90          & 25.60          & 25.30          \\ \cline{2-2} \cline{4-7} 
                                    & 20                                &                                                                           & 26.30          & 26.60          & 25.80          & 25.65          \\ \hline
\multirow{3}{*}{JSMA}               & 10                                & \multirow{3}{*}{\textbf{0.90}}                                            & 30.05          & 30.15          & 27.95          & 27.25          \\ \cline{2-2} \cline{4-7} 
                                    & 15                                &                                                                           & 29.60          & 28.45          & 25.80          & 28.00          \\ \cline{2-2} \cline{4-7} 
                                    & 20                                &                                                                           & 30.25          & 29.40          & \textbf{31.10} & 30.60          \\ \hline
\multirow{3}{*}{C\&W ($l_2$)}        & 10                                & \multirow{3}{*}{\textbf{13.20}}                                           & 54.15          & 53.35          & 52.55          & 52.00          \\ \cline{2-2} \cline{4-7} 
                                    & 15                                &                                                                           & 53.70          & 53.55          & 49.55          & 50.40          \\ \cline{2-2} \cline{4-7} 
                                    & 20                                &                                                                           & 53.10          & \textbf{54.95} & 50.35          & 52.85          \\ \hline
\end{tabular}
}\\
\end{table}

\noindent \textbf{Results on Tiny ImageNet:} The results are reported in Table \ref{resuts-tiny-prop}. The DeepFool attack reduces the accuracy from $66.60$\% to $23.60$\%, but the proposed defense model boosts the performance up to $52.00$\%. The C\&W $l_2$ attack is considered as one of the complicated optimization-based attacks, and researchers suggested that the defense algorithms must be evaluated against such attacks \cite{carlini2019evaluating}. In the proposed defense, when $20$ perturbations with $k=2.0$ are generated for training, the performance on C\&W $l_2$ images improved by $75.97$\%. Similarly, for the JSMA attack, which only perturbed the most salient pixels of an image, the proposed defense is able to improve the performance by $97.11$\%. The effectiveness of the proposed algorithm across multiple attacks and databases shows its strength in handling adversarial data.

\subsection{Comparison with Adversarial Training}

To further showcase the strength of the proposed defense algorithm, a comparison with the existing defense algorithm based on adversarial training is performed. In this research, adversarial training is performed using the adversarial examples generated using DeepFool, FGSM, and JSMA attacks. For adversarial training, the adversarial examples generated corresponding to the original model are used to further fine-tune the original model. The results are reported in Table \ref{resuts-comp-prop}. Each of the adversarially trained models is evaluated against each of the attacks implemented and results are compared with the proposed defense model. It is observed that after adversarial training, the performance of the original model trained with adversarial examples increased to $72.61$\% - $84.01$\% against all attacks. On DeepFool and FGSM attack, the adversarially trained model using FGSM ($\epsilon=2$) yields the best classification accuracy in comparison to other adversarially trained models. The best adversarially trained accuracy is $0.99$\% and $2.88$\% lower than the proposed defense model. On the JSMA attack, best recognition performance is obtained when the original model is fine-tuned using the JSMA adversarial images itself. On a strong C\&W attack, the best adversarially trained model is obtained using the adversarial examples generated using another strong attack i.e., DeepFool. The accuracy of the proposed defense model is $2.02$\% higher than the DeepFool trained model. This shows that the proposed model is robust against any attack without even requiring the knowledge of the attacks. However, for adversarial training, the knowledge of the attack and the samples generated from the attack is required for adversarial training which may not be available in real-world scenarios.

\begin{table}[t]
\centering
\caption{Comparing the performance of the proposed algorithm with adversarial training (AT) for different attacks in terms of classification accuracy (\%) on the CIFAR-10 database.}

\label{resuts-comp-prop}
\setkeys{Gin}{keepaspectratio}
\resizebox*{0.475\textwidth}{0.475\textheight} {
\begin{tabular}{|l|c|c|c|c|}\hline
\multirow{2}{*}{\textbf{Model}}   & \multicolumn{4}{c|}{\textbf{Testing Attacks}}                                                        \\ \cline{2-5}
                         & \multicolumn{1}{l|}{\textbf{DeepFool}} & \multicolumn{1}{l|}{\textbf{FGSM}} & \multicolumn{1}{l|}{\textbf{JSMA}} & \multicolumn{1}{l|}{\textbf{C\&W ($l_2$)}} \\ \hline
AT with DeepFool & 82.50                        & 81.53                    & 74.27   & 84.01                 \\ \hline
AT with FGSM ($\epsilon=2$)     & 83.06                        & 82.54                    & 73.51     &  81.87             \\ \hline
AT with FGSM ($\epsilon=8$)     & 79.02                        & 79.53                    & 73.19      &  79.80            \\ \hline
AT with FGSM ($\epsilon=16$)     & 79.90                        & 79.44                    & 72.61       & 79.99            \\ \hline
AT with JSMA     & 81.09                        & 80.10                    & \textbf{75.63}             &  79.70     \\ \hline
\textbf{Proposed}                 & \textbf{85.94}                        & \textbf{83.53}                    & 73.88 & \textbf{86.03}                   \\ \hline
\end{tabular}
}
\end{table}

\subsection{White-Box Experiments}

The white-box setting refers to the condition where an attacker has complete knowledge of the model, including the defense strategy. We have attacked the strongest defense model trained on original clean samples using $k=2.0$ for CIFAR-10 (given in Table \ref{resuts-cifar10-prop}) and $k=3.0$ for MNIST database (given in Table \ref{resuts-mnist-prop}). Inspired from literature \cite{madry2017towards} and to make the comparison with recent works \cite{mustafa2019adversarial,jang2019adversarial,bai2019hilbert}, we have performed the attack using one of the complex attacks, i.e., PGD attack. The perturbation size is set to $\epsilon=0.3$, while the step size varies from $\epsilon/4$ to $\epsilon/10$ for the MNIST database. On the CIFAR-10 database, the different $\epsilon$ values such as $2/255$, $8/255$, and $16/255$ with varying step size ranging from $\epsilon/4$ to $\epsilon/10$ is used to craft an attack. 
On comparing the performance of the proposed defense model and the existing algorithm L2L-DA on the MNIST database, it is observed that the proposed defense model is performing better by a small margin.  However, a significant difference in the performance of the proposed defense model and the existing algorithm L2L-DA is observed on the CIFAR-10 database. The performance of the proposed algorithm is atleast $29.84$\% better than L2L-DA \cite{jang2019adversarial}. In the experimental settings of the other two recent algorithms, on the CIFAR-10 database, the performance of the proposed algorithm is $36.49$\% and $5.96$\% better than \cite{mustafa2019adversarial,bai2019hilbert}, respectively. On the other hand, the proposed algorithm is $58.94$\% better than \cite{mustafa2019adversarial} on the MNIST.


\subsection{Computational Complexity}

Adversarial training is a two-step process, i.e., generation of adversarial samples, which require learning of perturbation for each example and training of a model with adversarial images. Let $n$ be the total number of original images, and an attack algorithm takes $m$ time/steps to generate/learn an adversarial image corresponding to each original image. Thus, it takes $O(m.n)$ time to create all the adversarial samples. Let us assume that the $O(k)$ time is made by the model to train on $n$ samples after that, the total time taken for the complete process is $O(m.n + k)$. In the proposed algorithm, the perturbations are sampled from the distribution and used for the model's training. Let $c$ be the number of perturbations sampled from the Gaussian distribution corresponding to each original image. Let us assume that for sampling operation corresponding to each original image, a constant time is taken.  Thus, the total time for sampling corresponding to $n$ real images is $O(n)$. The size of the database becomes $c.n$, and the time required to train the model is $O(c.k)$. Since $c$ is a constant, we can ignore this, and the time becomes asymptotically equivalent to $O(k)$. Thus, the entire process's total time for the entire process is $O(n+k)$, which has a smaller functional growth than $O(m.n + k)$.

\section{Conclusion}


The most effective defense algorithms in the literature are based on retraining of the target model using the adversarial examples. However, computing specific adversarial perturbations to retrain the network might not reduce sensitivity against other adversarial noise. To design a better defense model, it is important to unravel the behavior of attack algorithms against different databases and define a certain bound within which the defense model is robust against both seen and unseen attacks. In this research, we define a bound within which the visual appearance of the image is preserved while performing adversarial manipulation and term it as a \textit{visual imperceptible bound} (VIB). Using this bound, we propose a measure to evaluate the effectiveness of the attack algorithms on different databases. Further, we propose a defense algorithm to train a model that outputs the same prediction for the input samples present within the visual imperceptible bound. On evaluating the performance of the proposed defense algorithm, it is observed that it is robust against both seen and unseen attack algorithms. Further, the proposed defense is independent of the target model and adversarial attacks, and therefore can be utilized against any attack. 

\section*{Acknowledgment}

A. Agarwal is partly supported by the Visvesvaraya PhD Fellowship. R. Singh and M. Vatsa are partially supported through a research grant from MeitY, India. M. Vatsa is also partially supported through the Swarnajayanti Fellowship by the  Government of India.

\small{
\bibliographystyle{ieee_fullname.bst}
\bibliography{egbib}
}

\end{document}